# When to Intervene: Detecting Abnormal Mood using Everyday Smartphone Conversations

John Gideon, *Member, IEEE,* Katie Matton, Steve Anderau, Melvin G McInnis,
and Emily Mower Provost, *Senior Member, IEEE,*

**Abstract**—Bipolar disorder (BPD) is a chronic mental illness characterized by extreme mood and energy changes from mania to depression. These changes drive behaviors that often lead to devastating personal or social consequences. BPD is managed clinically with regular interactions with care providers, who assess mood, energy levels, and the form and content of speech. Recent work has proposed smartphones for monitoring mood using speech. However, these works do not predict when to intervene. Predicting when to intervene is challenging because there is not a single measure that is relevant for every person: different individuals may have different levels of symptom severity considered typical. Additionally, this typical mood, or baseline, may change over time, making a single symptom threshold insufficient. This work presents an innovative approach that expands clinical mood monitoring to predict when interventions are necessary using an anomaly detection framework, which we call *Temporal Normalization*. We first validate the model using a dataset annotated for clinical interventions and then incorporate this method in a deep learning framework to predict mood anomalies from natural, unstructured, telephone speech data. The combination of these approaches provides a framework to enable real-world speech-focused mood monitoring.

✦

## 1 INTRODUCTION

BIPOLAR disorder (BPD) is a mental illness characterized by pathological swings of mood, energy, and emotional states. Illness states may be along a continuum ranging from a paralyzing depression to uncontrolled mania. Affected individuals are impacted with impaired psychomotor functioning, emotional dysregulation, and altered cognitive processes [1]. BPD affects up to 4% of Americans, when all forms are considered [2]. Diagnosis, management, and monitoring of BPD is clinically based and relies on regular interactive assessments with care providers. Early detection of emerging symptoms and problems among people with BPD significantly improves outcomes. However, simply increasing the level of personal clinical monitoring is time consuming and financially impractical [3]. Recent work has proposed automatic mood monitoring as a solution to this dilemma, as it could provide ongoing care in a cost-effective manner [4], [5].

Speech monitoring is a promising candidate, as speech is influenced by the mood fluctuations of BPD and reflects the level of the psychomotor activity in the nervous system. Expression patterns and content of speech are governed by an individual's prevailing mood and emotional state. Depressed speech typically has lower energy, and is slow and monotonic, while manic speech is faster, louder, and higher energy when compared with healthy speech [6]. Changes in pitch, energy, and rhythm are typically linked to changes in mood [7], [8], [9], [10]. Estimates of emotion [11], [12] and patterns of language use [5], [13] have also been shown to be useful measures of mood are the focus of this work.

Research has shown that fatigue with medical monitoring devices emerges over time, particularly with self-report and strategies requiring interactive use [14]. However, prior work has generally relied on speech collected in a manner requiring active participation [10], [13], [15], [16] or in laboratory settings [17], [18]. In order for a system to be used for long-term and continuous mood monitoring, it must be unobtrusive and integrate into the daily life of individuals. Collecting speech samples from daily smartphone use is becoming increasingly popular [4], [5], [10], [13], [15], [16], [17], [18], [19]. But most studies still maintain certain restrictions on the data, including:

1) **Recording environment** – Data collections limited to noise-reduced laboratory settings [17], [18]
2) **Task** – Requires a repetitive task with an app [13], [15]
3) **Conversation type** – Data limited to subject calls with study clinicians [10], [16], [17], [18]
4) **Required within-subject labelled training samples** – Methods necessitate a potentially lengthy enrollment period for new subjects [16], [19]

Previous efforts have focused on predicting the intensity of mania and depression symptoms. However, symptom severity thresholds alone are insufficient. Different individuals may have different levels of symptom severity considered healthy, which we refer to as their baseline, and this baseline may change over time. More nuanced information is needed in order to make clinical adjustments or interventions for disease management [20]. This may include the specific characteristics of an individual [21].

PRIORI (Predicting Individual Outcomes for Rapid Intervention) is a passive mood monitoring system that predicts the need for intervention using natural phone conversations at the University of Michigan [10], [22]. Figure 1 gives a high-level system diagram of the PRIORI app and data analysis pipeline. In the data analyzed in this study, subjects were provided with a smartphone and app that recorded their side

• *J. Gideon, K. Matton, S. Anderau, M.G. McInnis, and E. Mower Provost are with the University of Michigan, Ann Arbor, MI, USA.*
*Email: {gideonjn, katiemat, standera, mmcinnis, emilykmp}@umich.edu*

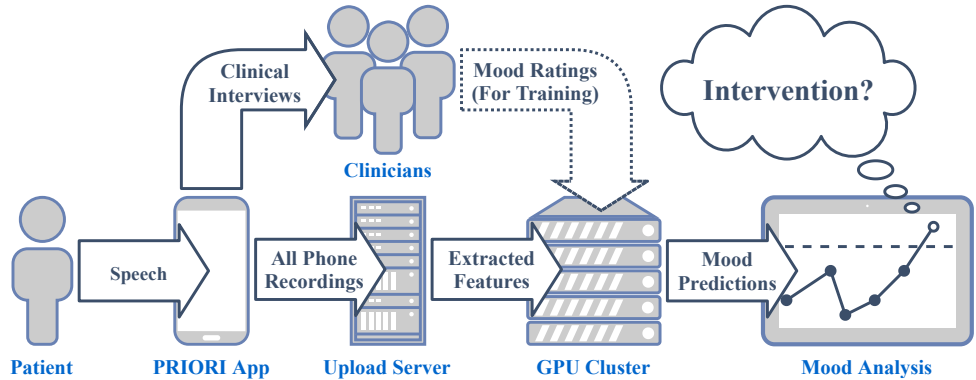

Fig. 1. The PRIORI system. Subjects use the PRIORI app to automatically record their speech as well as engage in phone-based clinical interviews. The app uploads speech to a server for feature extraction. These features and the mood ratings from the clinical interviews are used to train a mood prediction system. In this work, we explore how to determine which mood states are anomalous and potentially merit an intervention.

of all telephone conversations that they made or received. This collection resulted in two distinct types of data: (1) assessment calls - interactions between the subjects and study clinicians that generated the symptom severity labels and (2) personal calls - all other data. Symptom severity is measured during the assessment calls using the Young Mania Rating Scale (YMRS) [23] and the Hamilton Depression Rating Scale (HDRS) [24]. All data are uploaded to our servers for analysis and symptom severity prediction. While our prior work has stopped at this point [5], [10], [16], [22], this work goes further and predicts the need for clinical intervention from the recorded speech.

Intervention prediction requires intervention labels, or knowledge of when a clinician would choose to act. We obtained these labels by annotating a subset of the PRIORI corpus, referred to as the PRIORI Annotated Mood dataset (PRAM). These labels were generated by clinical chart review and annotation. We found that clinicians typically identified interventions based on symptom severity ratings that were abnormally high compared to an *individual's* baseline mood.

The PRAM labels were used to create an intervention detection system that could personalize over time, using techniques from the anomaly detection literature, which we refer to as *Temporal Normalization*, or TempNorm. TempNorm initializes with a baseline (a description that captures the range of typical behavior) for the general user population. As the system receives information from an individual, it first transforms these ratings into a continuous value indicative of the abnormality of the symptom severity. It then updates the baseline to personalize to the patterns *of each individual*. We validate the TempNorm framework on the intervention dataset. In particular, we investigate the trade-off between a conservative model that slowly adapts to each subject's baseline and one that instead reacts more strongly to recent mood. We show that TempNorm can be used to transform the symptom severity ratings to effectively predict if an intervention should occur. TempNorm significantly improves on a system using only a single population threshold, achieving an unweighted average recall (UAR) of 0.93±0.04, versus a UAR of 0.80±0.15.

We next investigated the ability of our system to automatically predict interventions from speech, rather than the clinician-assessed symptom severity measures. We combined a neural network with a middle layer consisting of TempNorm. We achieved a UAR of 0.70±0.14 and 0.68±0.12 for clinical and personal conversations, respectively. We find that transcript features perform best for the clinical calls, most likely due to their structured format, while both transcript and emotion features work well for natural speech. These results establish the first results for detecting interventions using clinically-collected and, critically, unstructured natural speech.

The novelty of this paper includes: (1) An outcome-based annotation of bipolar mood that identifies the need for clinical interventions; (2) The first work framing bipolar mood intervention detection in the context of anomaly detection using TempNorm; (3) The detection of anomalous mood solely from unstructured, natural speech, enabling real-world applications.

## 2 RELATED WORKS
### 2.1 Methods for Speech Mood Recognition

Depression is associated with speech that is lacking energy, monotonic, slowed, and poorly focused [6]. It has been measured in the acoustics of speech using changes in the amplitude, pitch, energy, shimmer, jitter, zero crossing rate and harmonic to noise ratio [7], [8], [9], [25]. Additionally, speech and silence timing, vocalization time, pause variability, speaking rate, and the amounts of short pauses have been shown to be effective at depression detection [25], [26], [27]. Other work has demonstrated that estimates of emotion could be used to improve depression detection [11], [12].

Manic speech is characterized as rapid, louder, less coherent, and more difficult to interrupt (pressured speech) [6]. Previous work has measured mania using rhythm, coherence, and the amounts of short pauses in speech [27], [28]. Additionally, increased pitch has been proposed as a sign of mania [7], [8]. Carrillo et al. identified a relationship between emotional intensity and manic mood using transcribed interviews [29]. Huang et al. differentiated BPD from unipolar depression by detecting the presence of manic speech using an attention-based Long Short-Term Memory (LSTM) classifier [30].

Research has shown that mood recognition can be improved by adapting a system to subject data, allowing it to pick up on subject-specific symptomatology [16]. However, this requires labeled data for all subjects, making the scaling of such systems difficult. Another approach from affect prediction is to use speaker normalization to reduce differences between subjects [31], [32], [33]. While this doesn't require labeled data, it often breaks causality by using all data to calculate the normalization parameters, thus using future subject data in forming the predictions of earlier samples. In a practical system, this would require an enrollment period to determine normalization parameters before being able to make predictions. Enrollment may also need to be carefully constrained to certain types of speech (e.g., non-symptomatic) or else the system may learn incorrect parameters.

While previous work has been essential to the understanding of speech changes in BPD, it often relies on speech collected in a manner not necessarily representative of natural speech. Several emerging efforts address this. Grünerbl et al. developed a phone app to recognize bipolar mood states using subject-dependent modelling [19]. Huang et al. focused on the detection of depression by asking subjects to record their speech using an app in natural environments [15]. Despite differences in noise characteristics due to device and environment, they were able to detect depressed speech using both short utterances [15] and landmark bigrams [13]. Pan et al. performed similar experiments for mania by using an app to record a patient's phone call with a psychiatrist [17], [18]. The call was conducted in a noise-suppressed laboratory environment, with the call being conversational in tone. Our previous work on BPD has demonstrated depressed and manic mood detection in passively recorded phone calls [10], [16]. We show that mood recognition (especially mania) is improved by addressing variability in clinical recordings due to device differences [10]. Our work has also shown the benefit of adapting a depression recognition system to subjects, using a hybrid population-general and subject-specific system [16].

Fundamentally, these works all still require some sort of active participation from subjects. One of the challenges in mobile health engagement is *app fatigue* – individuals tire of interactions with programs over time or ignore evaluations [14], [34]. Passive techniques are likely to be more successful in longer-term monitoring. These techniques should also be able to work in a variety of environments to facilitate continuous monitoring. Faurholt-Jepson et al. demonstrated the feasibility of detecting bipolar mood in everyday phone conversations [4]. Matton et al. reported transcript-based features extracted with Automatic Speech Recognition (ASR) are indicative of depression in BPD using natural speech [5]. However, these and other methods have not shown that interventions could be automatically triggered. This limits their use in intervention-driven applications, since the predictions are not directly related to clinical action. The output and evaluation of such systems should instead match their proposed clinical use [35].

Researchers have used reinforcement learning (RL) to explore this concept of interpretable and actionable systems for interventions. Typical supervised machine learning requires pairing predictions with ground truth labels. However, this is not always feasible when there is an unclear relationship between single actions and meaningful outcomes. RL instead defines and optimizes for long-term goals using domain-specific *reward functions*. For example, work in epilepsy has been evaluated using a reward function that penalizes the occurrence of seizures and learns the optimal pattern of deep brain stimulation [36]. Research in sepsis has employed a reward function tied to patient mortality and proposes the ideal personalized clinical intervention strategy [37]. Work in HIV has used a reward function based on changes in blood test measures and selected the best combination of drugs for therapy [38]. Our work differs from that in RL in that our ground truth directly pairs each sample of speech with an indication of the need for intervention.

## 2.2 Anomaly Detection

Anomaly detection identifies unusual measures in data, with common applications including outlier removal and fraud detection [39]. Basic methods of anomaly detection can take advantage of the distribution of data and designate points above a certain standard deviation or other measure as anomalies. For example, different forms of the moving average (MA) and variance can be used to de-trend and scale sequential data, as in [40], [41], [42]. Autogressive (AR) models are commonly used in sequence anomaly detection to forecast the likely value of the next sample using a certain number of prior samples [43]. Anomalies can then be designated as a deviation of an actual data-point from the predicted value. Autoregressive-moving-average models, or ARMA, combine both the MA and AR models, and are effective at detecting anomalies in a variety of fields [44], [45], [46]. All of these models may take into account prior knowledge about the domain, such as seasonal trends and and the base rate of occurrence of anomalies.

Recent work has employed neural networks to detect anomalies in data with unknown distributions. Autoencoders have been used to learn a compressed representation of the data, with anomalies identified by higher reconstruction error [47]. Malhotra et al. trained an LSTM to forecast time series predictions and then classified deviations from actual values as anomalies [48]. Generative adversarial networks, or GANs, can learn a latent space where anomalous data can then be more clearly distinguished [49]. These methods are unsupervised and focus on capturing aspects of the input with higher variance. Similar approaches have been ineffective at representing affect, as emotion and mood have a much lower varying nature when compared to other aspects of speech [50], [51].

Supervised anomaly detection is typically trained using standard classification methods with one category for normal data and one for anomalous data [39]. The main difference from typical classification problems is that the label distribution is strongly unbalanced and biased against anomalies. This bias can result in being unable to learn a robust representation for anomalies, due to a lack of examples. Additionally, it may be difficult to find clear representative samples of anomalies [52]. This is because anomalies are not necessarily defined by the presence of certain attributes, but are instead defined by the amount of deviation from a typical distribution. Modelling the problem as regression instead of classification can help avoid this issue by defining a continuous label for the abnormality of each sample [39]. However, previous work has not yet examined mood monitoring in the context of anomaly detection, leaving the definition of mood abnormality undefined. Our work establishes this definition using the exponential moving average (EMA) and exponential moving variance (EMVar), similar to [40], [41], [42], to track typical mood for subjects. We then define mood abnormality as a mood rating's deviation from this continuously updated baseline, as discussed in Section 5.

## 3 DATA

The PRIORI Dataset is a collection of natural speech from smartphone conversations of individuals with BPD [10], [12], [16], [22]. The subjects were recruited from the HC Prechter Longitudinal Study of Bipolar Disorder at the University of Michigan [21], and focuses on a subset that includes 51 subjects with BPD type I or II and nine healthy controls.

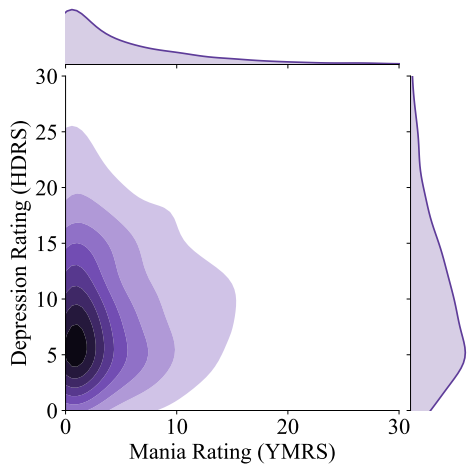

Fig. 2. The distribution of the mood ratings collected during the study.

Upon enrollment in PRIORI, each study subject was given an Android phone (Samsung Galaxy S3, S4, or S5) with the PRIORI app installed to passively record their end of all phone conversations made or received during the study on their study device. Subjects were asked to use this device as their primary phone for the duration of the study. The audio was encoded as 8 kHz WAV files and encrypted. The calls were then uploaded to a server for decryption and automatic analysis (described in later sections). Each subject was enrolled in the study for between six and 12 months (an average of 32±16 weeks). The dataset snapshot used in this work includes 51,970 phone calls, totaling 3,997 hours.

A small subset of the full PRIORI dataset was annotated for feature development. We first split subject recordings into 623,984 total segments. We then annotated a smaller portion of these with activation and valence, including 12 subjects with 13,611 segments (25.2 hours) of speech. This subset is known as the PRIORI Emotion dataset [12]. Additionally, we manually transcribed 25 hours of speech for use in transcript-based features.

Subjects were also asked to call a clinician to retrospectively rate their mood each week. These calls followed the interview format of the Young Mania Rating Scale (YMRS) [23] and the Hamilton Depression Rating Scale (HDRS) [24]. The subjects made these calls using their PRIORI study phone, recording the clinical interaction. We denote these recordings as *assessment calls* and all other non-clinical recordings that occur in the course of daily life as *personal calls*. Over the course of the study, 1,516 assessments were performed – 1,268 of which were recorded. This difference is due to either the subject not using their phone to conduct an assessment or issues with the app. Additionally, a smaller subset of subjects were asked to complete another YMRS and HDRS to rate their mood for just that day. We call these the *day-of mood ratings* and use only these data for intervention annotation. This work instead focuses on the week-retrospective ratings for analysis, as they are available for all subjects. Figure 2 shows the distribution of the YMRS and HDRS ratings.

Extensive clinical assessments are completed within the HC Prechter Longitudinal Study [21]. In addition, subjects provided permission to review medical records and both are included in the annotation process.

## 4 CLINICAL ANNOTATION

We created a new annotated subset of the PRIORI data, called the PRIORI Annotated Mood dataset (PRAM) to better understand when and why interventions are needed. We define *interventions* as changes in the treatment plan, such as emergency room visits, hospitalizations, or drug modifications to negative mood events. This definition corresponds to the concept of *necessary clinical medication adjustments* – an important measure of illness stability as reflected in the number of alterations in care considered necessary [53].

**Clinical Summary:** The University of Michigan electronic health record system was accessed to review relevant health information available for each subject during their time in PRIORI. This included all clinical notes and laboratory results from the Michigan Medical system, as well as other health systems that shared their records. Each subject's data was summarized to include relevant information about subject mood, clinical condition, or recent changes in either.

**Flagging Application:** We developed an application to review the clinical data for each week of participation in the PRIORI study and identify decision-making points for interventions (Figure 3). Annotation is divided into weeks based upon the date of each YMRS and HDRS retrospective interview. The ratings are displayed for each week, as well as the suicide subscore from the interview and day-of mood ratings, if available. The application allows annotators to browse all mood ratings and clinical summaries up to and including the day of the assessment, but prevents them from seeing future information. This ensures that annotators can only make decisions based on the data that would have been available at the time when a decision to intervene was made. The application allows the annotators to flag or not flag each week for intervention and then rate their confidence on a 1-3 scale (weak to strong). Annotators can indicate whether the intervention is urgent (needed within 24 hours) or a non-urgent follow-up. Once the annotation for a given week has been submitted, the application advances to the next week and previous submissions cannot be modified.

**Annotation:** Four clinically trained members (3 PhD and 1 MD) of the bipolar research team annotated all available clinical and research data. Each session consisted of a group of at least two annotators, who came to a consensus on the need for intervention. Groups were given summaries of clinical data before each subject's enrollment in PRIORI to establish a baseline summary of their medical history. Annotators were asked to complete subjects' entire annotations in one session to maintain consistency. Presently, 26 subjects totalling 555 weeks (71 with flagged interventions) have been annotated. Annotators were asked to provide reasons underlying the flagging to learn contributing factors that led to intervention recommendations. Common rationale used to flag for interventions include:

- High YMRS or HDRS, compared with personal baseline
- Lack of improvement from previous weeks
- Severity of clinical symptoms, e.g., suicidal thoughts

## 5 TEMPORAL NORMALIZATION

During clinical annotation, the most common rationale for flagging an intervention was mood ratings substantially above a subject's baseline. Because of this, we focus on how to model a subject's baseline using the YMRS and HDRS ratings to best predict anomalous mood and the need for interventions. In order to be successful, our system should be able to do the following: (1) estimate a baseline for subject mood, (2) predict anomalies based on a deviation from this baseline, (3) produce actionable predictions, even when little subject data has been seen. This section presents an example of how to model BPD mood, motivated by these goals and conversations with our clinical collaborators.

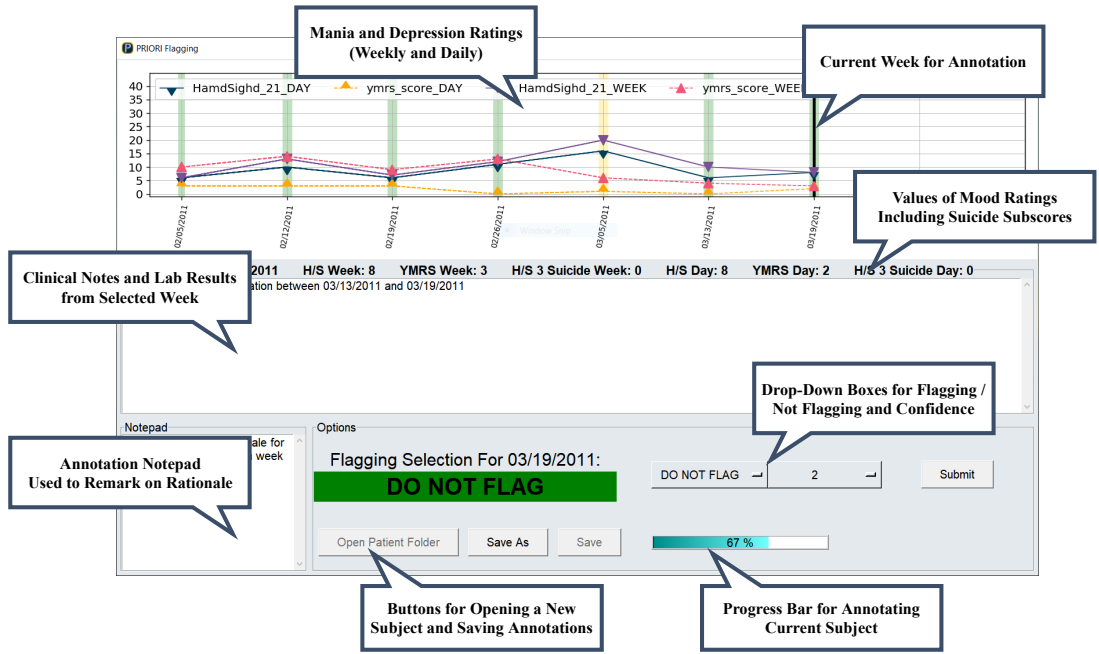

Fig. 3. The application used to read clinical data and flag interventions for each subject. The top of the application provides current and prior week-retrospective and day-of YMRS and HDRS ratings. The middle displays any clinical notes or lab results since the prior week. The user is able to click dates in the above graph to view notes for other weeks. The bottom is used to mark whether or not to flag a week for intervention and the confidence of the rating (1-3). The application advances to the next week upon submission and entries cannot be modified afterwards.

We formalize the problem as anomaly detection, using the EMA and EMVar, similar to [40], [41], [42]. This converts the problem from contextual anomaly detection to an easier point anomaly detection. For simplicity, we denote this procedure as *Temporal Normalization* (TempNorm) [1]. We ground TempNorm in its application to the YMRS and HDRS ratings. The full process is described in Algorithm 1 and shown in Figure 5, with each step explained in the following text.

We first subtract the YMRS and HDRS ratings by six and divide by four, based on an initial estimate of mean and standard deviation. These values were selected to closely match the within-subject rating means and standard deviations. This normalization also maps a rating of ten to one standard deviation from the mean. This is desirable, as ten was the threshold used in previous PRIORI experiments in the definition of a symptomatic state [10], [16]. Because the mood ratings should now have an approximate mean of zero and standard deviation of one, we initialize the EMA to zero and the EMVar to one. This establishes a subject's starting baseline as the population's baseline. We call this initial state the *population prior*.

We then approximate each subject's baseline over time using the EMA and EMVar. We normalize new data points by subtracting the EMA and dividing by the EMVar before updating these running statistics using the new values. This

---

1. An interactive demonstration of Temporal Normalization is available at http://www.johngideon.me/projects/TempNorm/

**Algorithm 1** Temporal Normalization (TempNorm)

**Input:** $X$, the 1d array to be normalized
**Input:** $t_{1/2}$, the half-life parameter
**Output:** $Y$, the 1d normalized output array
1: $\lambda \leftarrow 1.0 - \sqrt[t_{1/2}]{0.5}$  ▷ Get decay from half-life
2: $\mu \leftarrow 0$  ▷ Initialize EMA to 0
3: $\sigma^2 \leftarrow 1$  ▷ Initialize EMVar to 1
4: **for** $i = 1, ..., length(X)$ **do**  ▷ Loop through all samples
5: $\quad \Delta \leftarrow X[i] - \mu$  ▷ Get sample and EMA delta
6: $\quad Y[i] \leftarrow \Delta / \sigma$  ▷ Normalize current sample
7: $\quad \beta \leftarrow \lambda \times \Delta$  ▷ Scale delta based on decay
8: $\quad \mu \leftarrow \mu + \beta$  ▷ Update EMA
9: $\quad \sigma^2 \leftarrow (1.0 - \lambda) \times (\sigma^2 + (\beta \times \Delta))$  ▷ Update EMVar
10: **end for**

results in the first sample being unchanged because the original population prior is EMA=0 and EMVar=1. As the system sees new samples, the mood baseline and subsequent normalization adapts to that subject's patterns. TempNorm does not require the data to be sampled at a fixed rate, which is beneficial as subjects periodically have missing clinical assessments.

TempNorm requires one parameter, half-life ($t_{1/2}$), in order to control the contribution of new data to the running mean and variance. Half-life is described in units of the number of new samples needed to diminish the weight of old data to 50% (Figure 4). As half-life increases, the baseline takes longer to adapt to subject mood and remains conservatively closer to the original population prior. A half-life of infinity results in a system that only relies on the population prior and does not adapt to subject mood ratings. This makes it comparable to a system without TempNorm – one that has a single threshold across all subjects. Conversely, decreasing the half-life results in a system that concentrates more on recent data. A half-life of zero would only use the newest sample to calculate mean and variance. This would always result in a baseline variance of zero and produce divide-by-zero numerical issues in the normalization. Because of this, we restrict half-life to be any value greater than zero. Much of this work is focused on investigating the impact of half-life, as the model is heavily affected by its choice. We determine the weight of new samples, $\lambda$, using the following equation:

$$\lambda = 1.0 - \sqrt[t_{1/2}]{0.5}$$

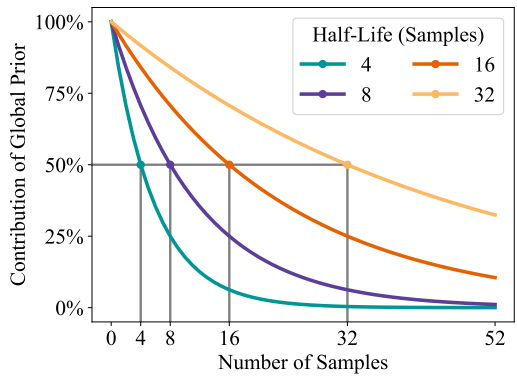

Fig. 4. The contribution of the original population prior distribution for different half-lives after certain numbers of samples have been observed.

TABLE 1
TempNorm Mood Ratings Compared with Annotation. Ratings below one are *typical*; those above two are *anomalies*; ratings between one and two are *unused*. The number of samples flagged for intervention in each region is shown in parentheses. Relying only on the global prior ($t_{1/2} = \infty$) results in a system with many false positives. Highlighted results are not significantly different from one another.

| $t_{1/2}$ | Number Subjects | Number Samples | | | UAR Mean ± Std. |
|---|---|---|---|---|---|
| | | Typical | Unused | Anomaly | |
| 1 | 13 | 322 (22) | 94 (24) | 136 (25) | 0.72 ± 0.19 |
| 2 | 11 | 352 (20) | 113 (26) | 87 (25) | 0.83 ± 0.13 |
| 4 | 11 | 366 (16) | 116 (28) | 70 (27) | 0.87 ± 0.12 |
| 8 | 12 | 339 (11) | 117 (25) | 71 (35) | 0.90 ± 0.07 |
| 16 | 13 | 317 (6) | 126 (18) | 84 (47) | 0.93 ± 0.04 |
| 32 | 13 | 282 (2) | 146 (16) | 99 (53) | 0.91 ± 0.07 |
| 64 | 13 | 268 (2) | 131 (5) | 128 (64) | 0.89 ± 0.09 |
| ∞ | 13 | 263 (2) | 71 (1) | 193 (68) | 0.80 ± 0.15 |

The EMA and EMVar are updated with each new sample, weighted by $\lambda$. This causes a hybrid global/speaker normalization of the data, depending on the half-life parameter and how many samples have been observed. This has the desired quality of gradually adapting to the most recent subject data, while also providing a measure of how anomalous a sample is versus the baseline.

The mood ratings are normalized separately, as the intensity of each subject's baseline mania and depression may differ. Furthermore, different subjects may have different correlations between these ratings. So for simplicity, we assume that both mood ratings are independent and model them with two separate TempNorms. This assumption will be later validated using the PRAM dataset.

We select a lower threshold of one standard deviation to represent a subject's typical mood. This matches the threshold for symptomatic mood used in our previous work [10], [16]. TempNormed YMRS and HDRS below this threshold are considered typical. We define an upper limit of two standard deviations as atypical mood, based on conversations with our clinical team. We define a *mood anomaly* as a sample with a TempNormed YMRS or HDRS above the upper limit. We leave the range between one and two standard deviations undefined, focusing only on regions with clear behaviors as in [54]. This work presents just one example of how to model BPD mood anomalies. Future work will investigate alternative models.

We validate our model using the PRAM annotations. We hypothesize that weeks flagged for intervention are weeks that our model should designate as anomalous (over two standard deviations from the EMA), while weeks marked without an intervention should be typical (within one standard deviation of the EMA). We do not assess model performance in the undefined region. It is important to note that the anomalous and typical category changes with half-life: TempNorm transforms the mood ratings based on this value (see Table 1).

We evaluate performance using unweighted average recall (UAR), an average of the recall over each category. This is desirable, since the distribution of the annotations is biased toward non-interventions. We sweep through different half-lives to gain insight into how annotators balance historical and new mood symptom information. We only consider subjects that have at least one typical and one anomalous week so that a valid UAR can be calcuated. As a result, the number of available subjects varies with half-life (see Table 1).

We fit a linear mixed-effect (LME) model in R [55], [56] to determine if the UAR of different half-lives are significantly different. We treat half-life as the fixed effect and subject as the random effect. All tests use a 0.05 significance threshold. We perform an analysis of variance (ANOVA) over the LME model to determine if there is any significant effect of half-life. We then perform a post-hoc pairwise comparison test with Tukey weighting using the emmeans package [57].

Our findings in Table 1 indicate that a half-life of 16 provides the best match to clinical annotation, with 0.93 ± 0.04 UAR, although multiple half-lives achieve comparable performance. We highlight half-lives that are not significantly different from one another. The rows that are not highlighted (1 and infinity) are significantly worse than at least one of the highlighted rows.

We show the full TempNorm procedure for two subjects using the half-life of 16 in Figure 5. Figure 5a shows a subject on which TempNorm performs well, while Figure 5b presents a difficult case (the subject in Figure 5b has particularly unstable mood). Figure 5a demonstrates how TempNorm can detect mood anomalies even given an increasing depression baseline (the turquoise lines). Blue markers outside the shaded regions indicate false positives, while red markers in the middle-most region are false negatives. The top plot shows the population prior system, which only relies on a fixed threshold and does not adapt to the increasing baseline. This results in 12 false positives, while TempNorm (shown at the bottom) decreases this to just four. Figure 5b gives an example of a subject with highly fluctuating mood and many intervention flags. The system normalizes many weeks to between one and two standard deviations because of the frequency with which this subject experiences heightened mood. This highlights the importance of clinical judgment in determining an intervention threshold – with this individual potentially benefiting from a lower cutoff. Despite this, TempNorm makes distinguishing extreme examples easier. The population prior with a fixed threshold results in eight false positive and two false negatives. TempNorm removes all false positives at the expense of two additional false negatives, but with a large increase in uncertain mood predictions (between one and two standard deviations).

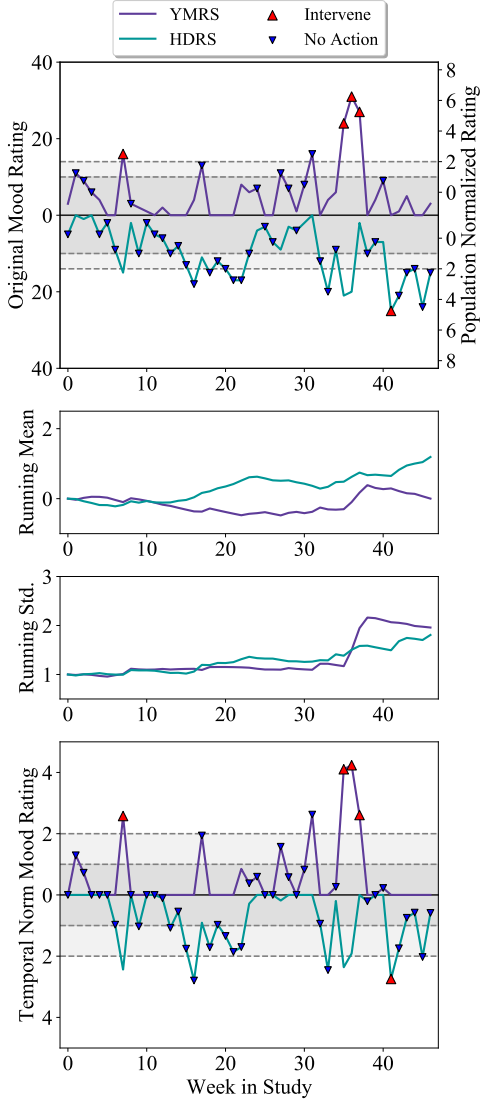
(a) Subject with good TempNorm performance

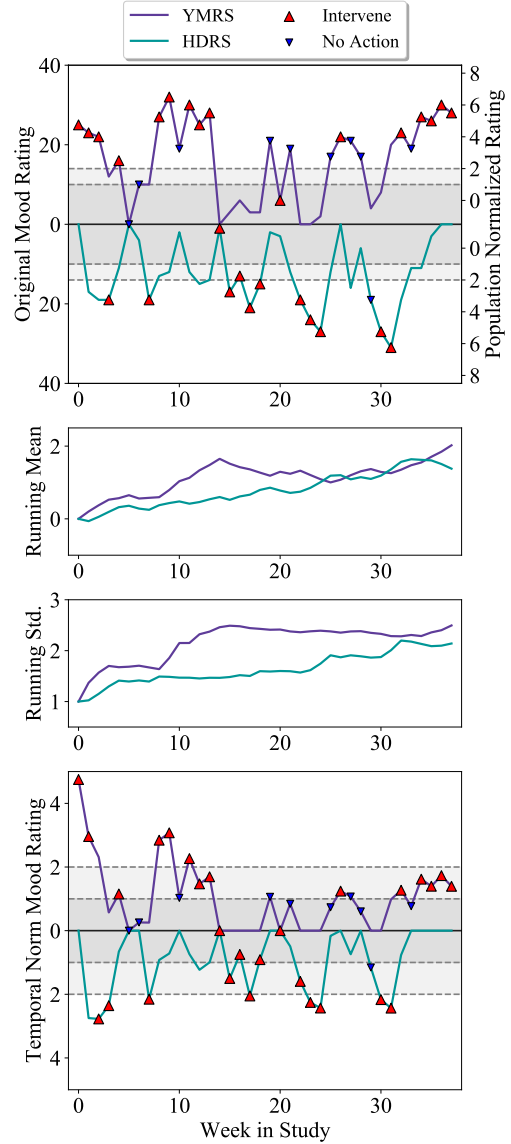
(b) Subject with difficult ratings for TempNorm

Fig. 5. TempNorm using a half-life of 16 samples for two subjects. Each gives four plots. (1) Depicts the original mood ratings and flags for intervention on the maximum of the mania or depression rating. The right y-axis gives the initial population normalized mood. The dashed lines and shaded regions depict the upper and lower mood thresholds of one and two standard deviations, respectively. These are used to differentiate typical and anomalous mood. (2) Gives the running scaled mean mood rating. (3) Gives the running scaled standard deviation of the mood rating. (4) Shows the TempNorm output with similar thresholds to the first plot. Normalized mood ratings below zero are truncated to zero.

This section has demonstrated the first advantage of TempNorm – it transforms the ground truth to resemble anomaly detection and creates more actionable predictions. While this section introduced TempNorm in the context of the YMRS and HDRS mood ratings, it would be easy to extend the procedure to other sequential data. The remainder of the paper explores the other two main benefits of TempNorm: (1) It initially behaves as a hybrid global/speaker normalization. After sufficient data, depending on the selected half-life, it acts like speaker normalization, providing the performance benefits of speaker normalization, without a requiring an enrollment period. (2) Each subject's mood ratings are self-normalizing, removing individual biases and resulting in reduced biases between subjects. This balances the dataset, making the learning of both typical and anomalous mood more straightforward.

## 6 FEATURES AND PREPROCESSING

We now focus on predicting the need for intervention from speech. In particular, we are interested in two different types of experiments, predicting mood anomalies from: (1) recorded clinical assessment calls, or (2) personal calls (non-clinical) from the same day as each assessment. We denote these as the *assessment* and *day-of* experiments, respectively. Note that the assessment calls themselves are never included in the day-of experiments. In this work, we focus only on calls from the day of the assessment because we hypothesize that they are most associated with the assessment label. Previous research has demonstrated that recency bias affects retrospective recall, causing clinical ratings to be strongly impacted by the most recent events [58]. For each experiment, we use different combinations of two speech feature sets – emotion and transcript.

## 6.1 Emotion Features

We have previously shown that there is a connection between fluctuations in emotion and mood [12], [59]. In this study, we validate this hypothesis by extracting measures of emotion and relating statistics derived from these measures to the clinical mood measures. We estimate emotion from the recorded speech using a model that we developed in our prior work [60], which we referred to as Multiclass Adversarial Discriminative Domain Generalization (MADDoG, described below). MADDoG allows the model to learn emotion, while also finding a representation that is similar across different datasets.

One of the key advantages of MADDoG is its ability to incorporate multiple datasets to create a more robust representation of emotion. As such, we train MADDoG using three emotion datasets – PRIORI Emotion [12] (discussed in Section 3), IEMOCAP [61], and MSP-Improv [62]. IEMOCAP contains emotional scripted and improvisational interactions between ten actors and totals 12 hours [61]. The MSP-Improv dataset comprises nine hours of emotional speech between twelve actors [62]. It includes scripted and improvisational scenes, as well as the spontaneous speech occurring between scenes.

All three datasets are labelled with the dimensional emotion measures of activation and valence. We bin each dimension into three categories of either low, mid, or high emotion.

The system uses short segments of uninterrupted speech as input. We extract 40-dimensional log Mel Filter Banks (MFBs) for each segment using Kaldi [63]. MFBs are measures of the frequency components of speech and have been successfully used to detect affect [64]. The model consists of three parts: (1) The **Feature Encoder** takes the input MFBs and creates a segment-level representation using a convolutional neural network (CNN). (2) The **Emotion Classifier** consists of fully connected layers and is trained to recognize the three binned categories of emotion from the encoded representation. (3) The **Critic** also contains a set of fully connected layers that takes the encoded representation as input and has three outputs, indicative of the segment's membership in each dataset.

The model is trained over 30 epochs with two alternating steps per epoch: (1) All weights besides the Critic are frozen and the Critic is trained so that each output estimates the Wasserstein Distance [65] in a one-versus-all manner between datasets. (2) The Critic weights are frozen and the Feature Encoder and Emotion Classifier are trained to classify emotion and minimize the Wasserstein Distance predicted by the Critic. This has the effect of both learning emotion and finding a compromise representation between datasets by "meeting in the middle". Our previous work has shown that this results in a more generalized classifier of speech emotion that is able to work on unseen data [59]. For further information on the MADDoG model and training procedure, please refer to [60].

Using this MADDoG model, we extract features for the final mood analysis. These features consist of binned segment-level estimates of both activation and valence that represent emotion dynamics over each assessment or day. However, we are interested in ensuring that no emotion labels are used in the eventual test set. To accomplish this, we train six different MADDoG models. The first five models are trained and tested in a round-robin manner using five folds and produces test predictions for all labelled data. One additional model is trained with all the 13,611 emotion-annotated segments and used to predict the remaining unlabeled segments. This results in three activation bins and three valence bins per PRIORI segment, or six total dimensions.

We hypothesize that the distribution of emotion over the course of an assessment or day is indicative of mood. To quantify this, we first concatenate all segments over the course of an assessment or day, depending on the experiment. We then take 31 statistics across the segments, which we previously demonstrated were related to mood [59]. This results in a final 186-dimensional feature set. This includes the mean, standard deviation, skewness, kurtosis, minimum, maximum, and range of the emotion bin predictions. We extract different percentiles (1, 10, 25, 50, 75, 90, 99) and percentile differences (25-50, 50-75, 25-75, 10-90, 1-99). We perform linear regression on the segment emotion estimates and incorporate the fit parameters and error ($R^2$, mean error, MSE) as features. Finally, we determine the percentage of the binned predictions above various thresholds (10%, 25%, 50%, 75%, 90% of the range).

## 6.2 Transcript Features

We transcribe the calls using an ASR model, which was implemented in Kaldi, an open-source, freely available speech recognition toolkit [63]. The model was built following the 'nnet2' recipe and was trained on the Fisher English Corpus [66]. When tested on the transcribed subset of the PRIORI dataset, it obtained a word error rate of 39.7%. We recognize that this is high. However, our data consists of unconstrained, natural speech in the presence of noise, so we expect imperfect transcriptions. Our previous work showed that transcript-based features extracted over ASR output were useful [5].

We extract call-level features from the assessment transcripts to use in our assessment experiments. For our day-of experiments, we concatenate the transcripts of all *personal* calls made on the day-of an assessment for each subject and assessment date (this excludes assessment calls). We extract day-level features from these merged transcripts. This results in a 208-dimensional feature set, which we divide into five categories. See [5] for more details.

We employ the Linguistic Inquiry and Word Count (LIWC) tool [67], a psycholinguistic analysis resource used in previous work to detect mental health states [68], [69], to compute the percentage of words belonging to 63 different language categories. Some of these categories are measures of **semantic content** and are related to psychological constructs (e.g. affect, biological processes) and personal concerns (e.g. work, death). The other categories measure aspects of **linguistic style**; these include 18 part of speech (POS) categories, three verb tense categories, and swear word, non-fluency, and filler categories. We extract 22 supplemental measures of linguistic style, including five additional POS categories, five POS ratios (e.g. adjectives:verbs), and 12 measures of speech complexity and verbosity (e.g. mean words per speech segment).

We apply speech graph analysis, introduced by Mota et al. to quantify thought disturbances in individuals with mania and schizophrenia [70], [71], as our final means of measuring linguistic style. We form speech graphs by representing each unique word as a node and inserting an edge for every pair of words uttered consecutively within the same speech segment. We create three graphs from each transcript that: (1) use the words directly, (2) use the lemmatized form of each word, and

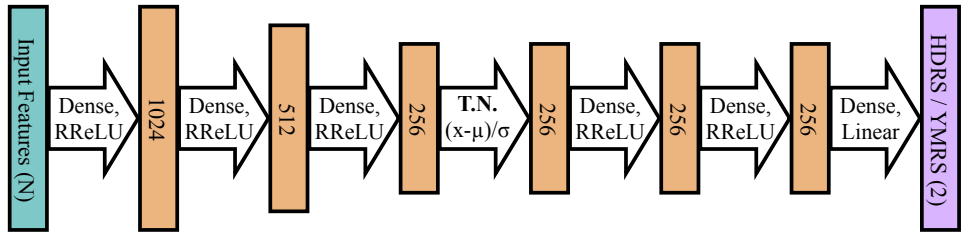

Fig. 6. The DNN used to predict mood abnormality, modified with a TempNorm Layer after the third hidden layer to learn a feature baseline.

TABLE 2
Restrictions causing the reduction of data for both the assessment and day-of experiments.

| Restriction | Number of Samples Assessment | Day-of |
|---|---|---|
| None | 1515 | 1515 |
| No healthy controls | 1319 | 1319 |
| Only S5 devices | 680 | 680 |
| 5 segments, 100 words, Valid ASR | 556 | 417 |
| Subjects need 8 samples | **533** | **369** |

(3) represent each word as its associated POS. We extract 12 measures per graph, including average degree, density, diameter, the size of connected components, and loop, node, and edge counts (see [70] for a full list). We also include a version of each feature that is normalized by total word count, providing us with 72 total graph measures.

We use Kaldi to generate aligned word and phone timing annotations for each transcript. From this output, we extract 43 features that quantify **speaker timing** patterns. We extract the same features for words, phones, and pauses: (1) statistics (mean, median, standard deviation, min, max) applied to the durations of all instances (e.g. mean word duration), (2) statistics (same set) applied to the per second timing within all segments (e.g. mean words per second across segments), (3) total count, and (4) per second timing over the whole transcript. We also extract total call duration, total subject speaking duration, ratio of subject speaking duration to total duration, total pause duration, ratio of pause duration to total duration, segment count, segments per minute, count of short utterances (lasting less than 1-second), and short utterances per minute, some of which were motivated by [72].

We use ASR confidence scores as measures of **speaker intelligibility** based on the idea that ASR has higher confidence for well enunciated speech. We apply statistics (mean, median, standard deviation, min) to the segment-level confidence scores to obtain four features (max was not used because it was almost always 1). Lastly, we quantify the presence of **non-verbal expressions** by extracting counts of instances of laughter and noise detected by the ASR model, normalized by word count.

### 6.3 Data Selection

We reduce our dataset to the highest quality data subset in order to focus on the impact of TempNorm on bipolar mood ratings (Table 2). We first remove all healthy controls to ensure the model specializes in individuals with BPD. We then remove subjects with phones other than the Samsung Galaxy S5s, as prior work has highlighted challenges with the other phone models [10]. We require data to have at least five segments, 100 words, and valid ASR transcripts (as in [5]). Finally, we require subjects to have at least eight samples (assessments/days), as this work focuses on adapting to a subject's baseline over time. This results in a total of 23 subjects and 533 samples (calls) for the assessment experiments and 17 subjects and 369 samples (days) for the day-of experiments. Note that removing data changes the ground truth because those samples no longer contribute to the baseline used in TempNorm (see Section 5).

## 7 MODELLING

The goal of the model is to predict the abnormality of the mood (TempNorm symptom severity) and to use this prediction to determine whether or not an intervention is needed. We use the same model and training methodology on the assessment and day-of experiments. We train two unimodal systems (i.e., only transcript and only emotion) and a multimodal early fusion system, resulting in a total of six systems, three for assessment and three for day-of.

We use a dense neural network (DNN), which has been effectively employed for mood recognition from static features [73]. It consists of six hidden layers with Randomized Leaky Rectified Linear Activation (RReLU) activations, as seen in Figure 6. The output layer has a linear activation and predicts Temporally Normalized YMRS and HDRS ratings.

We perform TempNorm in the feature space to match the label space. This makes both the labels and features relative to a subject baseline and allows for the detection of anomalies in both. Preliminary experiments without feature TempNorm had poor performance because the feature and label baselines drifted apart. We apply a *TempNorm Layer* to the 256-dimensional representation after the third fully connected layer. This applies TempNorm independently for each of the third layer's inputs using the same half-life parameter as the one used for the mood ratings. The feature model adapts to each subject's samples and, over time, performs subject-normalization over the mid-level representation. The model requires valid features to ensure that the label and feature EMA and EMVar remain synchronized. Future work will explore how to handle missing data (features or labels) and the placement of the TempNorm Layer.

The model is trained and tested in a round robin manner with five folds. Folds are kept subject independent by randomly assigning each subject to one fold. Data from each of the remaining subjects are split randomly with one fifth of samples used for development and the rest used for training.

Samples are batched by subject and are randomly reordered for data augmentation when training. Reordering the samples changes the baseline produced by the model and the ground truth at each time step, effectively increasing the amount of training data. Prior work has shown that mood changes in BPD are Markovian and that current mood is the most predictive of

TABLE 3
Assessment and day-of experiment results. The amount of subjects, typical (typ.) samples, and anomalous (anom.) samples is shown. Highlighted results show half-lives that do not produce significantly different results, given a certain feature set. An asterisk indicates results significantly better than the emotion features for the same half-life.

| $t_{1/2}$ | Total Number | | | Feature Set UAR (Mean ± Std.) | | | $t_{1/2}$ | Total Number | | | Feature Set UAR (Mean ± Std.) | | |
|---|---|---|---|---|---|---|---|---|---|---|---|---|---|
| | Subj. | Typ. | Anom. | Emotion | Transcript | Fusion | | Subj. | Typ. | Anom. | Emotion | Transcript | Fusion |
| 1 | 23 | 312 | 117 | 0.49 ± 0.09 | 0.61 ± 0.10* | 0.59 ± 0.09* | 1 | 17 | 216 | 86 | 0.53 ± 0.08 | 0.55 ± 0.08 | 0.56 ± 0.09 |
| 2 | 21 | 320 | 72 | 0.58 ± 0.14 | 0.67 ± 0.10* | 0.68 ± 0.12* | 2 | 17 | 231 | 55 | 0.59 ± 0.10 | 0.61 ± 0.11 | 0.63 ± 0.13* |
| 4 | 21 | 326 | 59 | 0.61 ± 0.12 | 0.68 ± 0.13* | 0.70 ± 0.14* | 4 | 17 | 223 | 49 | 0.59 ± 0.12 | 0.64 ± 0.13* | 0.65 ± 0.11* |
| 8 | 19 | 293 | 68 | 0.59 ± 0.11 | 0.66 ± 0.13* | 0.68 ± 0.12* | 8 | 15 | 192 | 43 | 0.63 ± 0.14 | 0.66 ± 0.12 | 0.68 ± 0.12* |
| 16 | 19 | 270 | 78 | 0.60 ± 0.09 | 0.68 ± 0.13* | 0.69 ± 0.14* | 16 | 15 | 176 | 61 | 0.62 ± 0.11 | 0.63 ± 0.12 | 0.65 ± 0.11 |
| 32 | 19 | 245 | 93 | 0.59 ± 0.12 | 0.68 ± 0.13* | 0.70 ± 0.14* | 32 | 15 | 166 | 76 | 0.57 ± 0.10 | 0.60 ± 0.12* | 0.61 ± 0.13* |
| 64 | 19 | 237 | 124 | 0.57 ± 0.10 | 0.67 ± 0.12* | 0.70 ± 0.15* | 64 | 15 | 159 | 96 | 0.54 ± 0.07 | 0.59 ± 0.13* | 0.59 ± 0.13* |
| ∞ | 19 | 223 | 174 | 0.54 ± 0.07 | 0.67 ± 0.12* | 0.68 ± 0.13* | ∞ | 16 | 163 | 127 | 0.51 ± 0.05 | 0.56 ± 0.11* | 0.56 ± 0.13* |

(a) Assessment Results (533 total assessment calls)     (b) Day-of Results (369 total days)

mood at the next assessment [74]. Due to the lack of long-term connection, we hypothesize that this reordering is an acceptable trade-off to augment the training data.

We fit the model using a weighted mean squared error (WMSE) loss to compensate for the relative rarity of anomalous mood. We calculate this WMSE loss for each mood output and sum them to form the total loss. Our model makes the assumption that our measure of mood abnormality approximately follows a standard normal distribution. We examine this assumption further in Section 9.2. Given this assumption, we set the weight of a sample to be inversely proportional to the probability density function (PDF) of the ground truth value. This gives higher weight to less common moods. We cap this weight at a maximum of 25 since the PDF increases rapidly for higher values, making rare samples dominate if unchecked. This method results in each sample having a different weight each epoch, depending on its order of appearance after randomization. This is because the abnormality (and weight) of samples is determined from the context of what comes before.

We train for 50 epochs. The first 10 epochs are used to pre-train the model without TempNorm. The next 40 epochs use TempNorm and validate to find the best stopping epoch for testing. The validation performance is also evaluated using WMSE. The WMSE takes the maximum of the two output mood predictions and the maximum of the TempNorm ground truth. This estimates the ability of the system to measure the abnormality of samples – focusing on the most extreme of the two moods. We found that this method of validation more closely matched the test setup and provided better performance.

Each subject's test performance is measured in UAR. We denote typical mood as TempNorm rating under one standard deviation, an anomaly as above two standard deviations, and unused as between one and two standard deviations. This strategy is similar to one by Georgiou et al., where they predict the amount of blame expressed in speech recordings of couple's therapy and just focus on the upper and lower 20% of samples [54]. The amount of prototypical data and available test subjects varies for each half-life, as discussed in Section 5.

It is important to note that although the test data we evaluate are only in the range of typical or anomalous, it is possible that our system will predict values in the unused range. We binarize the unused range to calculate UAR by introducing a threshold at 1.5 standard deviations. We assign estimates below 1.5 to typical mood and those above to anomalous mood.

For each of the two main experiments (assessment and day-of), we examine three combinations of features: (1) emotion, (2) transcript, (3) an early fusion of both. Our initial experiments found that applying global normalization (z-normalization using all training data) to transcript features worked best. No normalization was necessary for emotion. As such, we apply global normalization to the transcript features, in both the unimodal and early fusion sets. We run each experiment for 100 iterations, each with a different random seed. This helps to compensate for the randomness of selecting fold splits and the neural network initialization. This produces a final UAR matrix of size $100 \times \#Subjects$ for each if the six tested methods.

## 8 RESULTS

UAR measures our ability to judge the need for interventions. The performance is given as the UAR averaged over all subjects with enough data for the experiment (at least one typical and one anomalous sample). The number of subjects and samples changes depending on the half-life (see Section 5).

We test for the significance of each feature set and half-life on UAR by fitting a linear mixed-effect model, similar to Section 5. We consider feature set and half-life as fixed effects and subject and random seed as random effects. All tests use a 0.05 significance threshold. We first perform an analysis of variance (ANOVA) to determine if there is any significant effect of either feature set or half-life and find significance for both in each experiment. Next, we perform a post-hoc pairwise comparison, as in Section 5. We denote significantly better results than the emotion feature set with an asterisk in Table 3.

There were no significant differences between the transcript and fusion sets in either the clinical or the day-of calls. We describe the patterns in more detail in the following sections. We highlight the best performing half-lives for each feature set that are not significantly different from one another in Table 3.

### 8.1 Assessments

The assessment results are given in Table 3a. Regardless of half-life, the transcript and fusion feature sets significantly outperform emotion. There are no significant differences when appending the emotion features to transcript features (fusion). Assessment calls consist of the YMRS and HDRS interviews, and as such have a structure not present in natural speech. Because of this, the transcript features likely capture aspects of the questionnaires, giving them an advantage.

The performance of mood anomaly detection is mostly insensitive to half-life. The only exception is for a half-life of one. We find that a half-life of one changes the baseline too rapidly and results in significantly worse performance, using both emotion and fusion features. However, all half-lives between two and infinity provide stable results, and are not significantly different from one another. This stability is particularly evident in the transcript features. Again, this is likely due to the close relationship between the transcript features and the answers to the interview questions.

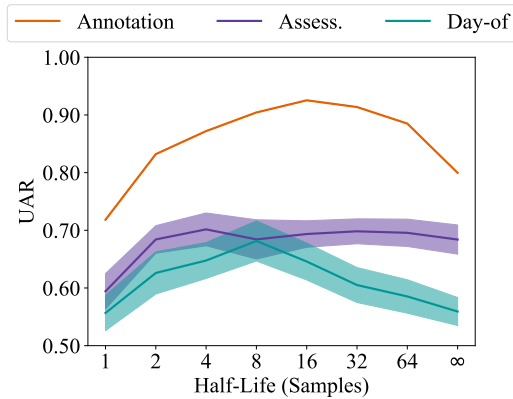

Fig. 7. Summary of experiment results. Assessment and day-of experiments use fusion features with the shaded region showing the standard deviation between random iterations.

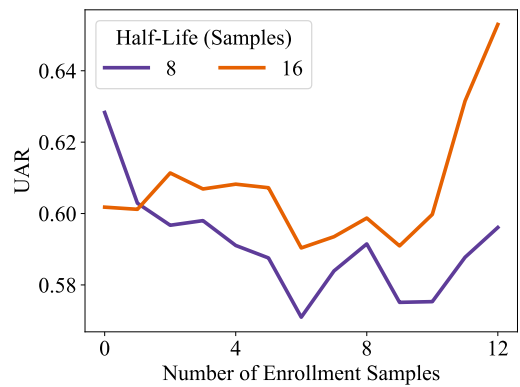

Fig. 8. The performance of the day-of fusion experiment, considering different enrollment periods and half-lives.

### 8.2 Day-of

The day-of results are given in Table 3b. The performance of the transcript features decreased, relative to the assessment experiment. This demonstrates how the efficacy of transcript features change when the assumptions of interview structure are no longer present – a result also shown recently in [5].

The emotion features attained similar performance in the two experiments. The similarity of the performance of emotion features between experiments indicates that they are capable of capturing mood-related aspects of speech present in both structured and natural conversations. In fact, the transcript features now only significantly outperform the emotion features in about half of the results.

We find that the day-of experiment is more sensitive to different half-lives, compared to the assessments. When working with non-clinical conversations it is especially important to establish a clear baseline. While a small half-life devalues old data too quickly, an overly large half-life takes too long to converge to subject normalization. In effect, the large half-lives result in systems that depend on the population prior distribution. This causes each subject's mood ratings to be biased, depending on how closely the population prior matches the subject's actual distribution. This mismatch of subject mood distributions can complicate classification and result in worse performance.

## 9 DISCUSSION

The choice of half-life has a strong effect on the outcome of most results in this work. Figure 7 shows the UAR from the annotation experiment, as well as the assessment and day-of experiments. Assessment and day-of experiments use the fusion features and the shaded error represents the standard deviation between iterations. Because there are no iterations for the annotation experiment, no error is shown. While the assessments are mostly insensitive to half-life (see Section 8.1), the half-lives of 8 and 16 provide the highest UAR for the day-of experiment and annotations, respectively. This section explores two factors that contribute to their performance: (1) enrollment length and (2) the distribution of the normalized ratings.

### 9.1 Enrollment Length

TempNorm begins with a population prior distribution and eventually learns a subject baseline. After sufficient samples it behaves like subject normalization, favoring recent data. Because of this, half-life controls both how quickly to disregard the population prior and the effective window length constructing a subject baseline. A half-life of 16 versus 8 will effectively incorporate double the samples into the subject baseline, but should also take about twice as long to converge.

We examine our day-of results after differing enrollment periods to see the change in performance during the adaptation process. For example, an enrollment period of four samples uses those samples to calculate the initial subject baseline, but not to make predictions. The remaining samples both continue to adapt the baseline and are the sole focus of the UAR calculation. We focus only on those seven subjects with enough data to consider an enrollment period of 12 weeks, so that the subjects stay the same with increasing enrollment.

Figure 8 shows how the mean subject UAR changes as the system accumulates subject data. The figure shows two half-lives, eight (purple) and 16 (orange). As before, we see that a half-life of eight provides the best performance when there is no enrollment data (enrollment of 0). However, it tends to worse performance when the subject baseline is initialized with enrollment data. Conversely, a half-life of 16 produces results that are fairly stable with enrollment periods of up to ten samples. This stability is likely associated with the broader effective window length for a half-life of 16. It then sharply increases in performance. While the method takes longer to achieve higher performance, the wider window provides a better subject baseline for comparison, versus a half-life of eight with no enrollment restriction. This demonstrates that higher half-lives can be beneficial – but only after sufficient data. Because there is not sufficient time for a half-life of 16 to reach its full potential, a half-life of eight provides better overall performance when considering no enrollment restrictions in the personal call experiment (see Table 3b).

### 9.2 Distribution of the Normalized Mood Ratings

While subject adaptation is one potential improvement caused by TempNorm, another is the self-normalization of the ground truth. Once a subject's baseline is determined, the mood is debiased and scaled so that it has a mean of zero and a standard deviation of one. As explained in Section 7, our model makes the assumption that the ground truth during training should have a unit normal distribution. Given this assumption, we weight each sample's loss inversely with respect to the unit normal PDF evaluated at the ground truth mood rating.

TABLE 4
Mean and Standard Deviation of Normalized Mood Ratings.

| $t_{1/2}$ | Mania | | | Depression | | |
|---|---|---|---|---|---|---|
| | Mean | Std. | $R^2$ | Mean | Std. | $R^2$ |
| 1 | 0.92 | 23.79 | 0.03 | 0.07 | 1.87 | 0.98 |
| 2 | 0.08 | 2.48 | 0.38 | 0.09 | 1.38 | 0.99 |
| 4 | -0.06 | 1.25 | 0.82 | 0.13 | 1.20 | 0.98 |
| 8 | -0.14 | 1.12 | 0.86 | 0.21 | 1.15 | 0.98 |
| 16 | -0.20 | 1.14 | 0.85 | 0.34 | 1.17 | 0.97 |
| 32 | -0.24 | 1.22 | 0.83 | 0.49 | 1.23 | 0.97 |
| 64 | -0.27 | 1.31 | 0.82 | 0.62 | 1.32 | 0.97 |
| $\infty$ | -0.28 | 1.57 | 0.78 | 0.96 | 1.69 | 0.94 |

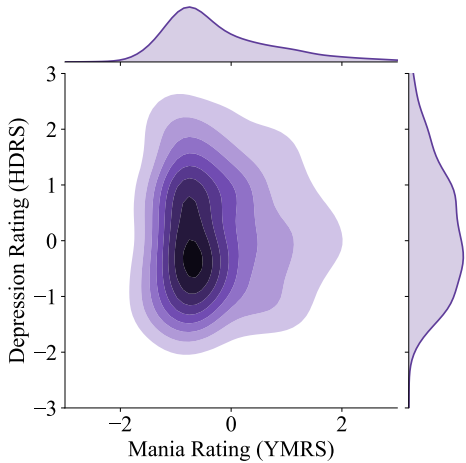

Fig. 9. Mood Rating Distribution After TempNorm ($t_{1/2}$ = 8).

In order to verify if this assumption is valid, we evaluated the distribution of the TempNorm mood at different half-lives. Table 4 shows the calculated means and standard deviations, as well as the $R^2$ value from a normal probability plot. We find that decreasing the half-live down to eight consistently causes the mean to trend toward zero, the standard deviation to trend to one, and the $R^2$ to get closer to one. However, this trend does not hold with very small half-lives. In particular, a half-life of one causes the standard deviation of normalized mania to increase sharply and the $R^2$ value to approach zero. Because the effective window of the EMVar becomes so small, only a few identical values in a row can cause it to approach zero. This is particularly a problem for mania because the distribution is skewed toward ratings of zero (Figure 2). After a few weeks with no mania symptoms, the next week with relatively higher mania symptoms will be substantially scaled upwards.

The compromise half-life between these two trends is at eight, whose normalized mood distribution is shown in Figure 9. This produces ratings that are self-normalizing, while still having a large enough window to prevent near-zero standard deviations when the original mood distribution is biased.

## 10 CONCLUSION

In this paper, we investigate not only how to estimate mood from natural speech, but also how to make meaningful predictions for each subject. In order to accomplish this, we collected the PRIORI Annotated Mood dataset – a set of annotations indicating when an intervention was needed, based on a variety of clinical factors. We then framed the problem of intervention detection as anomaly detection using TempNorm to transform mood ratings into a more actionable measure. This framework allows us to measure mood abnormality in clinical and natural speech using emotion and transcript features. Across all experiments, we determined that a half-life of eight or 16 provided the best compromise between subject baseline stability and adaptation. These half-lives allowed the model to learn a baseline quickly enough so that the mood ratings were transformed to a unit normal distribution – balancing the dataset and increasing classification performance.

The results of this study could form the basis for an intervention-driven clinical trial. TempNorm detects a continuous rating of mood abnormality and allows for variable thresholds at a personal level. For example, individuals experiencing app fatigue could have their anomaly threshold raised, while individuals with elevated mood instability may require a lower limit. Additionally, data from medical records may be used to more effectively initialize the system in a subject-specific manner. This would reflect actual clinical monitoring which takes into account any history to better target interventions.

Future work will address additional factors of variability to increase our usable data and make the system more versatile. For example, we excluded healthy controls from this work to focus on mood in individuals with BPD. However, future work will include healthy controls by allowing for a separate set of parameters. These will presumably be within the normal range of human behavior and should not require interventions. Additionally, we will use techniques to control for speech variability to allow for more devices and no longer limit our data to S5 recordings [10], [60]. We will also investigate other feature sets and incorporate other modalities, such as location data, to ensure that lack of speech does not result in an inability to detect mood anomalies. Finally, we will explore causal models to take advantage of any data occurring between mood ratings, in addition to day-of features. While previous work has shown a lack of causality for bipolar mood, this was on a week-to-week basis and may not apply to daily mood dynamics [74]. These improvements will allow us to increase the scope of PRIORI and get closer to an application capable of real-world use.

## ACKNOWLEDGMENTS

This work was supported by the National Science Foundation (CAREER-1651740). National Institute of Mental Health (R01MH108610, R34MH100404), the Heinz C Prechter Bipolar Research Fund, and the Richard Tam Foundation at the University of Michigan. Thanks to Ahmad Abu-Mohammad, Holli Bertram, Gary Graca, David Marshall, Bethany Navis, Kelly Ryan, Ariana Tart-Zelvin, and Kaela Van Til for their help with the annotations.

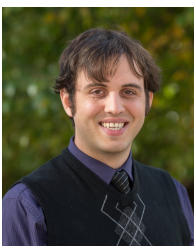

**John Gideon** is a Ph.D. candidate working with Professor Emily Mower Provost in Computer Science and Engineering at the University of Michigan, Ann Arbor. He received his B.S. in Electrical Engineering and M.S. in Computer Engineering from the University of Cincinnati, both in 2013. His research interests are the recognition of affect in speech and video for the improvement of medical care, as well as the design of multimodal assistive technologies for everyday tasks. He is driven by an underlying interest in human psychology and the way people perceive their interactions with one another. He is on track to graduate later this year and will then work with Toyota Research Institute on driver-facing technologies.

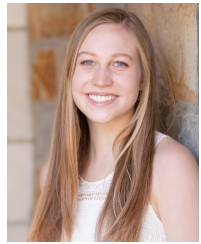

**Katie Matton** is currently pursuing her M.S. in Computer Science and Engineering at the University of Michigan, Ann Arbor. She received her B.S. in Computer Science and Engineering (summa cum laude) from the University of Michigan, Ann Arbor in 2018, where she was a member of the College of Engineering Undergraduate Honors Program. She works as a graduate student research assistant under the direction of Professor Emily Mower Provost. Her research interests are in affective computing, multimodal human-AI interaction, and speech and natural language processing. Her work is motivated by a belief in the potential of AI to improve the quality and accessibility of critical social services, such as mental healthcare, that are constrained by a lack of human resources.

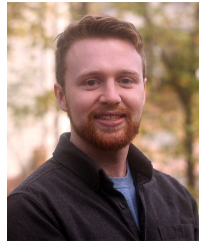

**Steve Anderau** is a Research Technician who joined the Michigan Medecine Heinz C. Prechter Bipolar Research Program in January 2019. He graduated from the University of Vermont (UVM) in May 2018 with a B.S. in mechanical engineering and a minor in computer science. Steve previously worked as a lead programmer and research engineer with the M-Sense Research Group at UVM, in collaboration with the University of Michigan, researching the use of speech pattern analysis in diagnosing young children at risk for anxiety disorders. He is currently part of the data management team on the PRIORI study here at the Prechter Program. Steve plans to pursue a higher education degree related to computer science and engineering in the coming years.

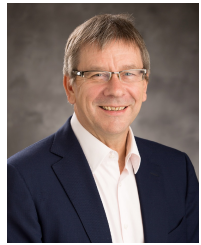

**Melvin G McInnis, MD,** is the Thomas B and Nancy Upjohn Woodworth Professor of Bipolar Disorder and Depression and Professor of Psychiatry. He is the Director of the HC Prechter Bipolar Program and Associate Director of the Depression Center at the University of Michigan. He is a Fellow of the Royal College of Psychiatry (UK) and Fellow of the American College of Neuropsychopharmacology. Dr. McInnis trained in Canada, Iceland, England, and USA, he began a faculty position in Psychiatry at Johns Hopkins University (1993) and was recruited to the University of Michigan in 2004. His research interests include the genetics of bipolar disorder and longitudinal outcome patterns in mood disorders. He has received awards recognizing excellence in bipolar research from the National Alliance for the Mentally Ill (NAMI) and National Alliance for Research in Schizophrenia and Affective Disorders (NARSAD). He has published over 250 manuscripts related to mood disorders research, and is widely engaged in collaborative research focused on identifying biological mechanisms of disease and predictive patterns of outcomes in mental health.

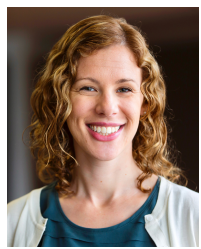

**Emily Mower Provost** is an Associate Professor in Computer Science and Engineering at the University of Michigan. She received her B.S. in Electrical Engineering from Tufts University, Boston, MA in 2004 and her M.S. and Ph.D. in Electrical Engineering from the University of Southern California (USC), Los Angeles, CA in 2007 and 2010, respectively. She is a member of Tau-Beta-Pi, Eta-Kappa-Nu, and a member of IEEE and ISCA. She has been awarded a National Science Foundation CAREER Award (2017), the Oscar Stern Award for Depression Research (2015), a National Science Foundation Graduate Research Fellowship (2004-2007). She is a co-author on the paper, "Say Cheese vs. Smile: Reducing Speech-Related Variability for Facial Emotion Recognition," winner of Best Student Paper at ACM Multimedia, 2014, and a co-author of the winner of the Classifier Sub-Challenge event at the Interspeech 2009 emotion challenge. Her research interests are in human-centered speech and video processing, multimodal interfaces design, and speech-based assistive technology. The goals of her research are motivated by the complexities of the perception and expression of human behavior.